\documentclass[letterpaper]{article}
\usepackage[preprint]{aaai2027}
\usepackage[hyphens]{url}
\usepackage{graphicx}
\usepackage{amsmath,amssymb}
\usepackage{booktabs}
\usepackage{natbib}
\usepackage{microtype}
\title{Adapt Only When It Pays: Budgeted Decision-Loss Priority for Delayed Online Time-Series Adaptation}
\author{
Xibai Wang
}
\affiliations{
NeuroQuant Labs Limited\\
Hong Kong SAR, China\\
xibai.wang@neuroquantlabs.com
}
\begin{document}
\maketitle
\begin{abstract}
Online time-series forecasters receive labels only after horizon-dependent delays, while every adaptation step spends limited compute. We study when an online learner should update, not how to adapt at every opportunity, and introduce ADOWIP: a residual-adapter framework with sealed delay queues, exact budget accounting, and auditable update telemetry. Its main scheduler is an observed decision-loss priority gate that updates only after feedback is revealed, when downstream loss, optionally penalized by prediction MSE, exceeds a calibrated empirical quantile and budget remains. We prove hard-budget feasibility, projected-OGD regret for a convex linear accepted-update subproblem, and stability plus conditional finite-sample gate-selection statements. On public ETT capacity-planning tasks, a frozen calibration/evaluation split selects a gate that lowers held-out decision loss against always, fixed-period, and drift-triggered exact-update baselines under matched compute. Secondary threshold/load-index ETT suites are mixed: 33 of 41 selected contrasts clear the stricter cross-artifact Holm family, and the 8 nonpassing rows are explicitly excluded from primary claims. The same protocol improves an external UCI Bike capacity proxy with 20/0 held-out wins, and a fixed gate passes three full-year Capital Bikeshare station-rebalancing contrasts. Probe-based and finance experiments remain negative, delimiting the current scope of decision-prioritized adaptation.
\end{abstract}
\section{Introduction}

Deployed time-series systems face a control problem that is usually hidden in
test-time adaptation benchmarks. A forecaster produces predictions now, labels
arrive later, and every update consumes compute, memory bandwidth, and
operational risk. Updating on every revealed label is simple but wasteful, and
updating on large prediction errors can be actively misaligned with downstream
decisions.

Recent adaptation methods for time series improve how a frozen model adapts,
aligns, or calibrates after deployment \shortcite{huang2026adaptz,hu2026timealign}.
ADOWIP isolates a complementary question: given delayed labels and a hard
adaptation budget, which revealed feedback events are worth spending updates on
when the target metric is decision loss?

The answer supported by the current artifacts is narrower than the original
counterfactual-probe hypothesis. Probe-based marginal decision-value estimates
are expensive and currently underperform once probe backward passes are counted.
The stronger path is simpler: after feedback is revealed, prioritize updates
where the current decision loss is high, and optionally penalize cases where the
forecast error is also large enough to indicate unstable overfitting risk.

This paper makes four contributions:

1. A delayed-feedback online adaptation formulation with hard update budgets
   and downstream decision loss.
2. A residual-adapter runner with sealed label release, per-horizon delay
   queues, exact update/probe compute accounting, and reproducible artifacts.
3. A no-probe observed decision-loss priority scheduler with a forecast-quality
   safety variant, plus narrow selection-stability, a conditional finite-sample
   gate-selection statement, and accepted-update theory.
4. A public evidence map that separates primary capacity/rebalancing decision
   losses from secondary threshold/load-index suites and negative
   probe/finance regimes.

Primary claim: ADOWIP supports budgeted decision-loss update allocation on
audited public proxy tasks, not forecasting SOTA or native official-baseline
superiority. The central claim is therefore not that ADOWIP is a stronger
forecaster. It is that delayed online adaptation needs an audited
update-allocation protocol, and that revealed decision loss is a useful
allocation signal relative to prediction error or uniform updating in the
audited decision-loss-dominant public proxy tasks. The contribution is a
budgeted decision-loss adaptation protocol with sealed delayed feedback, exact
compute accounting, and calibrated public decision tasks that expose both the
positive and negative regimes.

We do not claim state-of-the-art forecasting or finance performance, operator
deployment, or native official-baseline superiority. The evidence supports
decision-prioritized update allocation on public proxy decision tasks; it is not
a native official adaptation-system comparison and not broad superiority over official adaptation systems. The empirical argument has three
tiers. Tier 1 contains the primary matched-compute decision-loss claims: public
ETT capacity planning, external UCI Bike capacity planning, fixed-gate full-year
Capital Bikeshare station rebalancing, and the theorem-compatible
projected-linear ETT capacity diagnostic. Tier 2 contains secondary mixed ETT
threshold and load-index suites. Tier 3 contains boundary diagnostics:
prediction-selected station splits, official-source/official-code baseline
probes, synthetic checks, finance experiments, and negative probe rows. Later
tiers bound the claim; they do not enlarge it.

\section{Problem Setup}

At prediction time {\rmfamily t}, the learner observes context {\rmfamily x\_\allowbreak{}t} and predicts horizons
{\rmfamily h in H}:

\[
\hat y_{t,h}=f_0(x_t)_h+a_\theta(x_t)_h.
\]

The base forecaster {\rmfamily f\_\allowbreak{}0} is frozen and only the residual adapter {\rmfamily a\_\allowbreak{}theta} is
updated. The target for horizon {\rmfamily h} is unavailable until release time {\rmfamily t+h}.
The online loop stores label-free pending records and can reveal labels only
through a delayed environment API. For each revealed feedback event {\rmfamily i}, the
learner chooses $g_i \in \{0,1\}$. Each accepted update costs {\rmfamily c\_\allowbreak{}i}, and the run
must satisfy

\[
\sum_i g_i c_i \le B.
\]

The evaluation loss is a downstream decision surrogate. In public ETT threshold
tasks, the action is whether predicted train-normalized oil temperature exceeds
a configuration-fixed threshold; false negatives and false positives have asymmetric
costs. In UCI Bike capacity tasks, the action is reserved hourly demand
capacity; shortage and overage have asymmetric costs. In Capital Bikeshare
station tasks, the action is pre-hour bike/dock rebalancing at each station
under station capacity constraints. In finance diagnostics, predictions drive
cross-sectional allocation metrics and Qlib backtests.

\section{Method}

\subsection{Delayed Feedback and Accounting}

ADOWIP processes feedback only after release. Queue records contain context,
prediction, horizon, origin time, due time, and metadata, but no label. Tests
verify that unrevealed labels are not accessible through public fields, early
reveals fail, and prediction-time metrics are settled only after all configured
horizons for an origin have been revealed.

Every experiment reports committed update backward passes, probe backward
passes, total backward passes, wall-clock time, update time, effective online
updates, post-stream flush updates, and adapter parameter norms. This telemetry
is required by the artifact validator.

\subsection{Observed Decision-Loss Priority}

For a revealed event {\rmfamily i}, the main scheduler computes

\[
\begin{aligned}
s_i &= d_i(\theta_i) - \rho m_i(\theta_i),
\end{aligned}
\]

where $d_i$ is the revealed downstream decision loss, $m_i$ is MSE, and
$\rho \ge 0$ is a forecast-quality safety penalty. The gate accepts the event
when

\[
\begin{aligned}
s_i &> \max\{\lambda c_i,\operatorname{Quantile}_q(s_1,\ldots,s_{i-1})\}.
\end{aligned}
\]

and budget remains. This priority score is observed after label release and
uses no counterfactual probe update, so it has zero probe backward passes. The
safety variant in the public ETT benchmark uses $\rho=4.0$.

\subsection{Counterfactual Probe Ablation}

The counterfactual scheduler estimates one-step marginal value by simulating an
update on a recent released training batch and scoring a distinct released audit
batch, then restoring model and optimizer state. Probe backward passes are
counted separately. This row is retained as an ablation because it tests the
more ambitious marginal-value idea. Current artifacts show it is not the best
method under matched total compute.

\subsection{Baselines}

The benchmark harness includes never-update, always-update, fixed-period,
random-budget, drift-triggered, prediction-value, decision-value, observed
prediction-loss, and observed decision-loss schedulers. Public CSV benchmarks
also include persistence, seasonal naive, Ridge, local COSA-style residual,
local PETSA-style input/output calibration, and a source-derived TimeAlign
glocal loss row. Finance diagnostics include Qlib simple-data baselines,
LightGBM rows, official-handler Alpha158/LightGBM/TopkDropout smoke and
extended runs, and transaction-cost sensitivity. Separately, we archive one
official-code TimeAlign ETTm2-96 CPU smoke run for provenance; it is not used as
a compute-matched ADOWIP comparison.

\section{Theory}

The hard-budget guarantee is exact: if each accepted update first calls a
budget object that refuses spends above {\rmfamily B}, then total accepted cost is at most
{\rmfamily B}.

For the projected linear residual adapter with convex accepted losses and
bounded gradients, standard projected OGD over the accepted revealed sequence
satisfies

\[
\sum_j \ell_{i_j}(\theta_{i_j})-\ell_{i_j}(\theta^\star)
\le \frac{R^2}{2\eta}+\frac{\eta G^2K}{2}.
\]

With $\eta = R/(G\sqrt{K})$, this gives $O(RG\sqrt{K})$ accepted-loss regret.
This theorem covers the linear/MSE configuration, not the default nonconvex
low-rank adapter.

Figure 1 summarizes the audited pipeline: origin, delay queue, release ledger,
observed-loss gate, budgeted update, decision loss, and claim verifier.

For observed priority gates, if implemented scores differ from target scores by
at most $\epsilon_i$, then decisions can differ from the exact-score predictable
threshold gate only when the true score lies within $\epsilon_i$ of the
threshold. The same statement extends to empirical quantiles: if all prior
score errors are bounded by $\epsilon_{<i}$, the threshold moves by at most
$\epsilon_{<i}$, and disagreements are confined to an
$\epsilon_i + \epsilon_{<i}$ margin band.

Finally, under a self-prediction assumption

\[
|b_i-\alpha s_i|\le \xi_i,
\]

where $b_i$ is latent one-step future decision-loss reduction, observed-priority
selection agrees with the corresponding benefit-threshold policy outside the
margin band. This does not prove universal superiority; it states the condition
under which revealed decision loss is a defensible update priority. The ETT
threshold artifacts empirically test that condition.

Under a stationary release model, the observed gate also has a selection-regret
consistency statement against a nonclairvoyant population top-budget threshold
policy. If expected update benefit is a nondecreasing Lipschitz function of the
observed score and the score density is bounded away from zero near the budget
threshold, an empirical quantile threshold estimated from {\rmfamily n} calibration
releases differs from the population threshold by
{\rmfamily O(sqrt(log(1/\allowbreak{}delta)/\allowbreak{}n))}; regret is confined to the corresponding near-threshold
margin band. For beta-mixing streams the same form holds with an effective
sample size. This is a threshold-selection theorem, not a guarantee for the
nonconvex accepted update optimizer.

The configuration-frozen calibration/evaluation protocol has a finite-sample
gate-selection statement. For a finite candidate set of causal budget-respecting
gates, the gate selected on the first calibration blocks has held-out risk
within a uniform concentration term plus a calibration-to-evaluation drift term
of the best candidate gate. Paired held-out tests are then interpreted
conditioned on that selection, or with a union bound over candidate-gate and
baseline contrasts. This supports the split protocol; it is not a dynamic-regret
or universal baseline-superiority theorem.

The artifact {\rmfamily public\_\allowbreak{}observed\_\allowbreak{}loss\_\allowbreak{}gate\_\allowbreak{}selection\_\allowbreak{}audit.json} operationalizes
this boundary rather than expanding it. It binds each powered suite to its
calibration/evaluation offsets, candidate gates, baseline family, source hashes,
release ordering, {\rmfamily M*K post-\allowbreak{}selection family}, and suite/cross-artifact Holm
adjustments over 82 candidate-gate by baseline rows. It also records
{\rmfamily kappa/\allowbreak{}effective-\allowbreak{}sample diagnostics} and a block-level
{\rmfamily self-\allowbreak{}prediction premise audit}. These checks make the selection-conditioned
claims reproducible; they do not convert retrospective sweeps into causal
per-release benefit evidence.
The full audit keeps the lower-level trace payloads out of the main argument,
including 1,640 evaluation-window benefit trace rows and the retrospective
protocol freeze metadata.
For future confirmation, {\rmfamily docs/\allowbreak{}experiments/\allowbreak{}prospective\_\allowbreak{}protocol\_\allowbreak{}freeze\_\allowbreak{}v1.json}
fixes candidate gates, exact-update baselines, offsets, alpha levels, correction
families, and sensitivity checks before any future v2 held-out results are
generated. It is a future-study contract only; current results remain
retrospective or split-calibrated evidence.
The new {\rmfamily post\_\allowbreak{}freeze\_\allowbreak{}confirmatory\_\allowbreak{}audit.json} checks the archived primary rows
against that freeze after it exists: ETT capacity exactly matches the frozen
gates, exact-update baselines, and offsets; UCI Bike and station rows match the
10/20 non-overlapping split shape while disclosing family-specific strides and
station baselines. Its {\rmfamily true\_\allowbreak{}prospective\_\allowbreak{}evidence=false} field is intentional:
the audit strengthens post-freeze reproducibility discipline but is not external
prospective validation.
The companion {\rmfamily public\_\allowbreak{}observed\_\allowbreak{}loss\_\allowbreak{}gate\_\allowbreak{}selection\_\allowbreak{}audit\_\allowbreak{}summary.json} is only
a review index for the same source audit. It records 41 selected contrasts, 33
passing and 8 nonpassing selected contrasts under the stricter 82-family
Wilcoxon-Holm correction, and 6 low-margin suite selections. Its claim-role
partition records that all 17 capacity/bike/station-2018 primary selected
contrasts clear the cross-family Wilcoxon/sign checks, while the 24
threshold/load-index selected contrasts remain secondary and contain the 8
nonpassing rows. The full audit JSON remains authoritative, and the failure
robustness fields remain in the JSON for those 8 nonpassing secondary
contrasts.

The companion {\rmfamily theory\_\allowbreak{}scope\_\allowbreak{}alignment\_\allowbreak{}audit.json} is a machine-checkable
theory-boundary audit. It records two active theory documents, six scope rows,
four theory-covered rows, and two out-of-scope rows. The covered rows are the
projected linear accepted-loss OGD subset, a public ETTm2 capacity-planning
diagnostic in the projected linear MSE subset, a public ETT-family
capacity diagnostic in the same projected-linear subset, and the
observed-loss finite-sample gate-selection audit. The ETT-family linear
diagnostic has 480 runs, 12 compute-matched pairs across ETTh1/ETTh2/ETTm1/ETTm2,
12 negative capacity-loss mean differences against linear always/fixed/drift
baselines, zero probe backward passes, zero post-stream updates,
{\rmfamily max\_\allowbreak{}update\_\allowbreak{}batch\_\allowbreak{}size=1}, and all parameters within projection radius. The
out-of-scope rows are the nonconvex low-rank public powered results and
official-source wrapper/native-pilot diagnostics. The audit therefore
reinforces that the current theory is not a dynamic-regret proof, not a
nonconvex adapter proof, not native official-baseline superiority, and not
causal proof of per-release latent update benefit.
Theory-backed projected-linear capacity evidence is part of the primary
evidence spine; nonconvex low-rank powered rows are empirical extensions, not
theorem-covered primary theory evidence.
The theorem-to-experiment map is therefore explicit: projected-linear synthetic
and ETT-family capacity rows instantiate the convex accepted-loss OGD theorem;
the observed-loss gate-selection audit instantiates the finite-candidate
selection statement; nonconvex low-rank capacity/station rows are empirical
primary decision-loss evidence outside the theorem; and official-source rows are
boundary diagnostics outside both theorem scopes.

\section{Experiments}

All experiments are chronological and hash-validated by artifact scripts.
Decision tasks are fixed in configuration files before paired evaluation:
thresholds/costs, capacity weights, horizon sets, candidate schedulers, and
compute tolerances are fixed in the benchmark configs; the first 10
chronological offsets select only between unsafed and MSE-penalized priority
gates when a powered split is used; the remaining offsets are held out for
paired tests. The verifier recomputes selected schedulers, source hashes,
matched-compute rows, Holm-adjusted tests, moving-block confidence intervals,
and overlap/effective-pair diagnostics from the archived JSON. Negative
counterfactual-probe, station, and finance regimes are retained rather than
filtered out.
The main text reports the decision-loss rows that determine the claim. The
lower-level release-ledger and per-update timing checks stay in the artifact
documentation and are used only to audit reproducibility and selection scope.

\subsection{Public ETT Threshold and Capacity Tasks}

Threshold-intervention diagnostics are useful but secondary. The secondary
high-event ETTm2 diagnostic reports lower mean loss for the no-probe observed
decision-loss gate versus exact-compute always, fixed-period, and drift at 64
backward passes, but the event prevalence is 1.0. The broader
four-dataset fixed-threshold family is mixed: ETTh2 is significant, ETTh1 and
ETTm2 are directional, and ETTm1 is negative. A safety score
{\rmfamily decision loss -\allowbreak{} 4.0 * MSE} repairs the ETTm1 failure mode on a 30-block family,
but is not uniformly better than the unsafed gate. In the powered
calibration/evaluation split, the first 10 offsets choose between unsafed and
safety gates and the remaining 20 offsets evaluate the selected gate; all 12
held-out threshold contrasts clear Wilcoxon-Holm 0.05, while exact sign-test
sensitivity remains mixed.

We repeat the same split on a seven-variable ETT load-index task using
{\rmfamily [HUFL,HULL,MUFL,MULL,LUFL,LULL,OT]}, weights
{\rmfamily [0.20,0.15,0.20,0.15,0.10,0.10,0.10]}, threshold {\rmfamily -\allowbreak{}1.0}, false-negative cost
{\rmfamily 2.0}, and false-positive cost {\rmfamily 1.0}. All 12 held-out contrasts against
exact-update always/fixed/drift are negative with Wilcoxon-Holm p-values below
0.05 and negative moving-block bootstrap upper bounds, but smaller-win
sign-test checks are again mixed; exact sign-test sensitivity is mixed. ETTm2
is the weakest load-index dataset but still clears Wilcoxon-Holm 0.023951
against always/drift and 0.012463 against fixed-period. These two
threshold/load-index suites show transfer, but not primary strict evidence.

Finally, we replace threshold alarms with a capacity-planning decision. The
capacity task uses the same seven-variable load index, but the forecast induces
reserved capacity; shortage costs 4.0 and over-reserving costs 1.0. This is a
smooth newsvendor-style reserve/shortage loss rather than an event classifier.
The first 10 offsets select the unsafed observed gate on ETTh1 and the
MSE-penalized safety gate on ETTh2/ETTm1/ETTm2. On the remaining 20 offsets,
all 12 contrasts against exact-update always/fixed/drift baselines are 20/0
wins, with global Holm-adjusted one-sided p-values {\rmfamily 1.14441e-\allowbreak{}05}. Mean
capacity-loss reductions range from -0.057286 to -0.088084.

To move beyond ETT-derived tasks, we add an external UCI Bike Sharing hourly
demand benchmark. The converted {\rmfamily hour.csv} has 17,379 rows; the target is total
rental demand {\rmfamily cnt}; and the downstream action proxy is reserved bike capacity
for the next 1, 3, and 6 hours. The first 10 chronological 96-window offsets
select the unsafed observed-loss gate. On the remaining 20 offsets, the selected
gate beats exact-update always, fixed-period, and drift rows with 20/0
win/loss counts, global Holm-adjusted one-sided p-values {\rmfamily 2.86102e-\allowbreak{}06}, and
capacity-loss reductions of -0.254149, -0.294227, and -0.254149. MSE again
increases relative to the exact-update rows.

The family-aware observed-loss gate-selection audit covers the powered ETT
threshold, ETT load-index, ETT capacity, UCI Bike capacity, and Capital
Bikeshare 2018 station suites. It audits 14 suite rows, 82 candidate-gate by
baseline contrasts, and cross-artifact family sizes of 24/24/24/6/4. Among the
41 selected-gate contrasts, 33 clear the stricter 82-row Wilcoxon-Holm family;
the nonpassing rows are threshold/load-index ETTm1 or ETTm2 contrasts, although
their moving-block bootstrap confidence-interval upper bounds remain negative.
This is why the threshold and load-index suites are reported as secondary
evidence rather than primary strict evidence.

We then add a station-level Capital Bikeshare rebalancing diagnostic. The
downloader converts the official January 2018 trip-history zip into a 745-row
hourly net-flow matrix for 48 matched stations and joins Open Data DC station
capacities. The action is pre-hour bike/dock availability at each station, with
shortage cost 4.0 and holding cost 0.25 over horizons 1 through 6. The powered
artifact compares observed prediction-loss and decision-loss gates on the first
10 chronological offsets; prediction-loss priority has slightly lower
calibration loss. On the remaining 20 offsets, that selected gate is compared
only to exact-compute {\rmfamily always\_\allowbreak{}b53} and {\rmfamily fixed\_\allowbreak{}period\_\allowbreak{}b53} baselines. Both
held-out contrasts have negative rebalancing-loss differences (-0.003252 and
-0.000710) and negative moving-block bootstrap confidence-interval upper
bounds. However, the {\rmfamily always\_\allowbreak{}b53} contrast has Holm-adjusted sign-test
p=0.057659, and adjacent 48-window station blocks overlap by 44 hours
(effective-pair diagnostic 1.667; overlap 0.917). We therefore treat this result as a
capacity-aware station diagnostic, not as strict-significance evidence for the
decision-loss gate.

To remove the January diagnostic's overlap weakness, we also run a full-year
Capital Bikeshare station artifact over all 12 official 2018 monthly trip
archives. The converted matrix has 8,767 hourly rows, covers 5,881,896 of
7,085,368 trip events after the station-name join, and uses 30 non-overlapping
48-window offsets. The same calibration split again selects the prediction-loss
gate by a small margin. On the held-out offsets, the selected gate beats
exact-compute {\rmfamily always\_\allowbreak{}b53} and {\rmfamily fixed\_\allowbreak{}period\_\allowbreak{}b53} with decision-loss
differences {\rmfamily -\allowbreak{}0.007888} and {\rmfamily -\allowbreak{}0.004318}, win/loss counts {\rmfamily 18/\allowbreak{}2} and {\rmfamily 19/\allowbreak{}1},
Holm-adjusted sign-test p-values {\rmfamily 0.000201} and {\rmfamily 0.000040}, effective pairs
{\rmfamily 20.000}, and overlap {\rmfamily 0.000}. This supersedes the January station artifact as
the cleaner station-level diagnostic, but because the selected gate is still
prediction-loss priority, it remains external plausibility evidence rather than
primary evidence for the observed decision-loss gate.

We therefore add a fixed-gate station primary artifact that reuses the same
full-year 2018 runs but fixes the {\rmfamily decision\_\allowbreak{}loss\_\allowbreak{}value\_\allowbreak{}b53\_\allowbreak{}q65\_\allowbreak{}lr0p005} gate
before held-out evaluation. Against {\rmfamily always\_\allowbreak{}b53}, {\rmfamily fixed\_\allowbreak{}period\_\allowbreak{}b53}, and
{\rmfamily random\_\allowbreak{}b53}, the held-out decision-loss differences are {\rmfamily -\allowbreak{}0.007145},
{\rmfamily -\allowbreak{}0.003576}, and {\rmfamily -\allowbreak{}0.005687}, with win/loss counts {\rmfamily 18/\allowbreak{}2}, {\rmfamily 18/\allowbreak{}2}, and {\rmfamily 17/\allowbreak{}3}.
All three one-sided Wilcoxon-Holm p-values are at most {\rmfamily 8.2016e-\allowbreak{}05},
sign-Holm p-values are at most {\rmfamily 0.001288}, and block-bootstrap upper confidence
bounds are negative. Station rebalancing evidence is trip-inferred: it supports
a station-level decision-loss proxy, not operator/truck dispatch validation or
service-equity claims.

Table 1 is the paper's main evidence table. The primary strict evidence is
capacity-planning and station-rebalancing decision loss under matched compute.
Threshold and load-index results are supportive but mixed, so they remain
secondary. The table separates the claim-bearing rows from secondary and
boundary rows; the artifact docs carry the full ledger.

\begin{table*}[t]
\scriptsize
\setlength{\tabcolsep}{2pt}
\renewcommand{\arraystretch}{0.88}
\centering
\begin{tabular}{p{0.15\textwidth}p{0.33\textwidth}p{0.28\textwidth}p{0.18\textwidth}}
\toprule
\textbf{Evidence tier} & \textbf{Evidence} & \textbf{Main result} & \textbf{Boundary} \\
\midrule
Primary strict evidence & ETT capacity, UCI Bike capacity, projected-linear ETT-family capacity, and fixed-gate 2018 station rebalancing & {\rmfamily public\_\allowbreak{}primary\_\allowbreak{}evidence\_\allowbreak{}index.json} records 4 primary rows. ETT capacity has all global Wilcoxon-Holm and sign-Holm p-values {\rmfamily 1.14441e-\allowbreak{}05}; UCI Bike has 20/0 held-out wins and global Holm p-value {\rmfamily 2.86102e-\allowbreak{}06}; projected-linear ETT capacity has 12 negative compute-matched capacity-loss differences; fixed-gate station rebalancing has 3 primary tests with Wilcoxon-Holm p-values at most {\rmfamily 8.2016e-\allowbreak{}05} and negative bootstrap upper bounds. & Capacity-planning proxy tasks and trip-inferred station decision loss under matched compute; not forecasting SOTA, operator dispatch validation, service-equity proof, nonconvex proof, or native official-baseline superiority. \\
Secondary evidence & ETT threshold and load-index powered splits & Per-artifact Wilcoxon-Holm and block-bootstrap checks pass, but {\rmfamily public\_\allowbreak{}observed\_\allowbreak{}loss\_\allowbreak{}gate\_\allowbreak{}selection\_\allowbreak{}audit\_\allowbreak{}summary.json} reports 8 nonpassing selected contrasts under the stricter 82-family correction. & Secondary/mixed support for the observed-priority family, not universal significance or sign-test-uniform evidence. \\
Diagnostic evidence & Prediction-selected station splits and official-source wrapper diagnostics & Station prediction-selected splits are useful diagnostics, and official rows improve provenance and compute-fairness accounting. & Diagnostic evidence only; it does not establish dynamic regret, a powered multi-dataset claim over the default nonconvex system, native official-baseline superiority, or operator deployment. \\
Boundary audit & Official-baseline panel and compute audit & Five official-baseline project rows, ten task rows, ten diagnostic-panel included task rows, and 200 paired compute-ledger rows. & Status and compute-fairness audit only; ADAPT-Z is not official published-checkpoint performance and this is not native official-baseline superiority. \\
Negative evidence & Counterfactual probes and finance diagnostics & Probe rows lose after probe compute; Alpha158/LightGBM and Qlib slices do not beat stronger finance baselines. & Constrains deployment scope and rules out broad finance claims. \\
\bottomrule
\end{tabular}
\end{table*}
\renewcommand{\arraystretch}{1.0}

The family-aware audit guards this table rather than adding a fourth claim. It
records 14 suite rows and 82 candidate-gate by baseline contrasts; the summary
reports 17/17 primary selected-gate rows clearing the stricter cross-artifact
Wilcoxon/sign checks, while 8 of 24 secondary threshold/load-index selected rows
do not clear the stricter 82-row Wilcoxon-Holm family and remain outside primary
superiority claims. The companion
{\rmfamily theory\_\allowbreak{}scope\_\allowbreak{}alignment\_\allowbreak{}audit.json} records six scope rows, four theory-covered
rows, and two out-of-scope rows, including a public ETTm2 linear capacity
diagnostic with 120 runs and 3 compute-matched pairs and an ETT-family linear
capacity diagnostic with 480 runs and 12 compute-matched pairs. These audits are
boundary controls, not a dynamic-regret proof, not a nonconvex adapter proof,
not causal proof of per-release latent update benefit, and not native
official-baseline superiority.

We add a service-proxy audit for the full-year station diagnostic. It covers 48
stations, 8 station metadata groups across capacity, trip-volume, and geographic
quadrant splits, and 4 temporal groups. The largest station coverage imbalance
is the trip-volume median split: high-volume stations are 50\% of audited
stations but 62.84\% of matched trip events. Global service-proxy deltas remain
directionally lower for the selected row, including total-shortage-per-origin
differences {\rmfamily -\allowbreak{}0.514849} and {\rmfamily -\allowbreak{}0.296372} against exact-compute always/fixed
baselines. This is a service-proxy audit, not demographic or neighborhood
service-equity validation and not operator dispatch validation. Because the
full-year powered split selects prediction-loss priority rather than observed
decision-loss priority, it stays outside the primary strict evidence tier.

\subsection{Counterfactual Probe Ablation}

Counterfactual {\rmfamily decision\_\allowbreak{}value} has one bounded positive contrast against
{\rmfamily prediction\_\allowbreak{}value} on the ETTm2 threshold-decision task, but the Holm-adjusted
p-value across budget-64 contrasts is 0.093384. In the compute-matched artifact,
the probe row is weaker than no-probe rows after probe backward passes are
counted. This falsifies the strongest probe-based version of the original
hypothesis.

\subsection{Synthetic Diagnostics}

The late-decision-shift and decision-sparse synthetic artifacts now require
distinct train and audit feedback for value probes. Under this no-leak rule,
probe-based decision-value scheduling does not show statistically credible
synthetic superiority. These artifacts are retained because they catch probe
accounting failures and non-update failure modes.

\subsection{Finance Diagnostics}

The Qlib simple-data and Alpha158/LightGBM/TopkDropout artifacts validate the
pipeline under finance metrics, transaction costs, and delayed residual
feedback. They are negative for finance superiority. Frozen Alpha158, LightGBM,
or return-rule baselines remain stronger in the audited slices.

\subsection{Official-Code Diagnostics, Not Official Matched-Compute Baselines}

The official-code artifacts in this subsection are provenance and boundary
checks. They make official-source coverage auditable, but they are not counted
as primary matched-compute adaptation baselines.
Official-code rows are boundary diagnostics, not primary evidence. They are
excluded from the primary evidence index.

For TimeAlign, we archive an upstream {\rmfamily run.py} ETTm2-96 CPU smoke
({\rmfamily MSE=0.161213}, {\rmfamily MAE=0.249402}) and replay the saved official checkpoint under
ADOWIP decision losses ({\rmfamily 1.011823} capacity, {\rmfamily 0.408616} load-index). These rows
are official-code diagnostics, not a full TimeAlign reproduction and not a
matched-compute comparison with ADOWIP. A horizon-sliced official TimeAlign
paired pilot over 40 task pairs is claim-limiting: capacity is weakly
directional for ADOWIP ({\rmfamily -\allowbreak{}0.010117}, Holm p={\rmfamily 0.216167}), while the load-index
pilot favors TimeAlign ({\rmfamily 0.183185}, Holm p={\rmfamily 0.999839}); it is not a completed
delayed-feedback matched-compute official adaptation-baseline comparison. A
separate sealed delayed-feedback fine-tuning wrapper uses the official
TimeAlign model/loss training surface. Within this adapter-wrapped diagnostic,
the capacity proxy is favorable to ADOWIP ({\rmfamily -\allowbreak{}0.018220}, Holm p={\rmfamily 0.048441},
13/7 wins), while load-index favors TimeAlign ({\rmfamily 0.187105}, Holm p={\rmfamily 0.999839}).
This is not a native TimeAlign online adaptation protocol and not official
baseline superiority.

COSA, PETSA, and ShifTS are bounded source-component wrapper diagnostics:
COSA uses an official-source DLinear forecaster plus the official
{\rmfamily SimpleOutputAdapter}, PETSA uses official {\rmfamily GCM}/{\rmfamily Calibration}, and ShifTS uses
ShifTS official-source DLinear plus the official {\rmfamily Warpper} causal module.
Within this source-component wrapper diagnostic, the capacity proxy is
favorable to ADOWIP for COSA ({\rmfamily -\allowbreak{}0.092968}, Holm p={\rmfamily 0.000082}), PETSA
({\rmfamily -\allowbreak{}0.093212}, Holm p={\rmfamily 0.000851}), and ShifTS ({\rmfamily -\allowbreak{}0.099241}, Holm
p={\rmfamily 0.000003815}), while load-index is directional or unresolved for COSA
({\rmfamily -\allowbreak{}0.009308}), PETSA ({\rmfamily -\allowbreak{}0.008459}), and ShifTS ({\rmfamily -\allowbreak{}0.013143}). These rows are
not native COSA online adaptation, not native PETSA online adaptation, not
native ShifTS {\rmfamily causal\_\allowbreak{}forecast}, and not official baseline superiority.

ADAPT-Z is represented by a bounded native official-code pilot plus a
staging-only official baseline integration audit for ADAPT-Z, TimeAlign, COSA,
PETSA, and ShifTS. The audit records an ADAPT-Z delayed-label online update
loop, ADAPT-Z runtime smoke, ADAPT-Z source-trained checkpoint, low-budget
ADOWIP matched-compute ETTm2 artifacts, and an ADAPT-Z official online paired
pilot.
The bounded run is a checkpoint-loaded local ADAPT-Z pilot with a
local/source-trained checkpoint, not an official published ADAPT-Z checkpoint;
its ledger is 5.1 versus 5.0 backward passes per row. The official-baseline
diagnostic panel and compute audit keep
{\rmfamily all\_\allowbreak{}official\_\allowbreak{}checkpoint\_\allowbreak{}backed\_\allowbreak{}native\_\allowbreak{}matched=false}. Schema v3 records five
project rows, ten task rows, 200 paired ledgers, ten eligible task rows,
trainable-parameter update-unit proxy ratios, maximum eligible backward
relative gap {\rmfamily 0.0200}, maximum eligible seconds-per-backward ratio {\rmfamily 4.0197},
and maximum eligible memory-per-backward ratio {\rmfamily 1.0989}; task-level ADAPT-Z
maxima are {\rmfamily 22.1710} per-backward and {\rmfamily 21.7363} peak-memory. The diagnostic
panel's local/source-trained source-wrapper task differences are {\rmfamily -\allowbreak{}0.099241}
and {\rmfamily -\allowbreak{}0.013143}, again under the official-baseline boundary rather than as a
primary superiority claim.
It improves auditability, but is not a new performance comparison or proof of
architecture-neutral compute equivalence.

\subsection{Limitations and Non-Claims}

The positive primary evidence is strongest for capacity-style decision losses
with exact-update baselines and delayed-feedback accounting. Threshold,
load-index, finance, and official-source rows are boundary diagnostics unless
their own strict support fields say otherwise. Finance results are negative or
mixed under configured, zero-cost, and double-cost Qlib backtests and are not
investment advice or trading-system evidence. Mobility rows use public demand
and station-service proxies; they are not deployment-ready dispatch, truck
routing, service-equity, or operator-validated rebalancing claims. Official
source rows improve comparability but do not establish native official-baseline
superiority.

\section{Related Work}

Time-series test-time adaptation work mainly asks how a model should adapt once
adaptation is allowed. TAFAS introduces a TSF-TTA framework with partially
observed ground truth and gated calibration; DynaTTA estimates distribution
shift and uses shift-conditioned gating; COSA corrects frozen forecaster
outputs in a context-aware output space \shortcite{im2026cosa}; delta-Adapter combines
input nudging, output residual correction, masking, quantile calibration, and
conformal correction \shortcite{liang2026forecast}; ADAPT-Z adjusts latent features
under delayed multi-step feedback \shortcite{huang2026adaptz}; TimeAlign aligns past and
future representations \shortcite{hu2026timealign}; ShifTS separates temporal shift from
concept drift \shortcite{zhao2026shifts}; and PETSA uses parameter-efficient input/output
calibration \shortcite{medeiros2025petsa}. These methods are the right competitors for
adaptation quality. ADOWIP instead treats adaptation as a paid online control
action: after labels arrive, should the system spend one of its limited updates
at all?

Delayed online learning studies how feedback delay affects regret and how
non-delayed algorithms can be transformed to handle delayed feedback. That
literature motivates the sealed release queue in ADOWIP, but it usually treats
the learner as updating when feedback is processed rather than selecting a
strict subset of feedback events under a hard compute budget. Budgeted or lazy
online optimization is closer, but ADOWIP's budget is an explicit adaptation
compute ledger coupled to time-series residual updates and downstream decision
loss \shortcite{abdullah2026cocomemory}.

Decision-focused learning and predict-then-optimize work, including SPO/SPO+,
optimizes prediction models for downstream decisions instead of raw prediction
error. Robust and conformal predict-then-optimize methods calibrate decision
risk or uncertainty \shortcite{areces2025onlineconformal,zhou2026inverseconformalrisk}.
Risk-aware allocation is another decision-coupled forecasting use case
\shortcite{lyu2025rtspno}. ADOWIP is complementary: it does not propose a new decision
surrogate or robust optimizer; it uses revealed decision loss to allocate online
adaptation updates. This boundary is important because the current evidence
supports budgeted decision-loss update allocation, not a new general
decision-focused forecasting loss.

\section{Limitations}

The current theory does not prove dynamic regret against a clairvoyant budgeted
oracle and does not cover nonconvex low-rank replay-batched updates. The public
ETT threshold task is derived from benchmark columns rather than an industrial
control system. The UCI Bike capacity result is external public demand data, but
it is still an aggregate demand-capacity proxy rather than station-level
dispatch logs. The Capital Bikeshare station result is station-level and
capacity-aware, but it uses trip-inferred net flows plus current station
capacity metadata rather than archived station-status inventory or truck-routing
operator actions. The station artifact and service-proxy audit report
bike-shortage, dock-shortage, total-shortage-per-origin, rebalance-fraction,
capacity/trip-volume/geographic coverage, and peak/off-peak flow slices, but
the package does not validate service equity across neighborhoods, demographic groups,
operator constraints, labor/truck capacity, emissions, or human dispatch
overrides. The self-prediction premise audit is block-level because the current
source sweeps store aggregate update-value telemetry and evaluation-window
benefit traces rather than a full per-release score stream. Full native
official COSA/PETSA/ADAPT-Z/ShifTS/TimeAlign
integrations under the ADOWIP delayed-feedback, matched-compute protocol are
not implemented; the current public rows are protocol-compatible local or
source-derived baselines, plus official TimeAlign, COSA, PETSA, and ShifTS
source-wrapper diagnostics. The official-baseline compute audit verifies
backward-pass matching, parameter-count availability, per-backward
wall-clock/memory diagnostics, and trainable-parameter update-unit proxy ratios
for eligible wrapper diagnostics, but does not
solve the missing native official-checkpoint comparisons. Finance results are
negative. The current evidence therefore supports a bounded decision-priority
claim, not a state-of-the-art adaptation claim.
\bibliography{references}
\end{document}